\documentclass[11pt]{article}
\usepackage{coling2018}
\usepackage{times}
\usepackage{url}
\usepackage{epsfig}
\usepackage{amsmath}
\usepackage{amssymb}
\usepackage{subcaption}
\usepackage{graphicx}
\usepackage{bm}
\usepackage{soul}
\usepackage{xcolor}
\usepackage[group-separator={,},group-minimum-digits={3}]{siunitx}
\usepackage{multicol}

\newcommand{\fancyhl}[3][red]{\colorlet{#1#2}{#1!#2}\sethlcolor{#1#2}\hl{#3}}
\usepackage{CJKutf8}

\newcommand*{\affaddr}[1]{#1} 
\newcommand*{\affmark}[1][*]{\textsuperscript{#1}}
\newcommand*{\email}[1]{\texttt{#1}}

\title{SGM: Sequence Generation Model for Multi-Label Classification}

\author{Pengcheng Yang\affmark[1,2], Xu Sun\affmark[1,2], Wei Li\affmark[2], Shuming Ma\affmark[2], Wei Wu\affmark[2], Houfeng Wang\affmark[2]\\
\affaddr{\affmark[1]Deep Learning Lab, Beijing Institute of Big Data Research, Peking University}\\
\affaddr{\affmark[2]MOE Key Lab of Computational Linguistics, School of EECS, Peking University}\\
\email{\{yang\_pc, xusun, liweitj47, shumingma, wu.wei, wanghf\}@pku.edu.cn}\\
}

\date{}

\begin{document}
\maketitle
\begin{abstract}
Multi-label classification is an important yet challenging task in natural language processing. It is more complex than single-label classification in that the labels tend to be correlated. Existing methods tend to ignore the correlations between labels. Besides, different parts of the text can contribute differently to predicting different labels, which is not considered by existing models. In this paper, we propose to view the multi-label classification task as a sequence generation problem, and apply a sequence generation model with a novel decoder structure to solve it. Extensive experimental results show that our proposed methods outperform previous work by a substantial margin. Further analysis of experimental results demonstrates that the proposed methods not only capture the correlations between labels, but also select the most informative words automatically when predicting different labels.\footnote{The datasets and code are available at \url{https://github.com/lancopku/SGM}}
%
\end{abstract}

\section{Introduction}

\blfootnote{This work is licenced under a Creative Commons Attribution 4.0 International Licence. Licence details: \url{http://creativecommons.org/licenses/by/4.0/}}

Multi-label classification (MLC) is an important task in the field of natural language processing (NLP), which can be applied in many real-world scenarios, such as text categorization \cite{tc}, tag recommendation \cite{tr}, information retrieval \cite{ir}, and so on. The target of the MLC task is to assign multiple labels to each instance in the dataset. 

Binary relevance (BR) \cite{r1} is one of the earliest attempts to solve the MLC task by transforming the MLC task into multiple single-label classification problems. However, it neglects the correlations between labels. Classifier chains (CC) proposed by \newcite{ml_3} converts the MLC task into a chain of binary classification problems to model the correlations between labels. However, it is computationally expensive for large datasets. Other methods such as ML-DT \cite{ml_7}, Rank-SVM \cite{ml_8}, and ML-KNN \cite{ml_6} can only be used to capture the first or second order label correlations or are computationally intractable when high-order label correlations are considered.

In recent years, neural networks have achieved great success in the field of NLP. Some neural network models have also been applied in the MLC task and achieved important progress. For instance, fully connected neural network with pairwise ranking loss function is utilized in \newcite{r5}. \newcite{r4} propose to perform classification using the convolutional neural network (CNN). \newcite{r6} use CNN and recurrent neural network (RNN) to capture the semantic information of texts. However, they either neglect the correlations between labels or do not consider differences in the contributions of textual content when predicting labels. 

In this paper, inspired by the tremendous success of the sequence-to-sequence (Seq2Seq) model in machine translation \cite{machine_translation,luongPM15,sunxu}, abstractive summarization \cite{abstractive_summarization,globalencoding}, style transfer \cite{shen17,xu2018unpaired} and other domains, we propose a sequence generation model with a novel decoder structure to solve the MLC task. The proposed sequence generation model consists of an encoder and a decoder with the attention mechanism. The decoder uses an LSTM to generate labels sequentially, and predicts the next label based on its previously predicted labels. Therefore, the proposed model can consider the correlations between labels by processing label sequence dependencies through the LSTM structure.
Furthermore, the attention mechanism considers the contributions of different parts of text when the model predicts different labels. In addition, a novel decoder structure with global embedding is proposed to further improve the performance of the model by incorporating overall informative signals.

The contributions of this paper are listed as follows:
\begin{itemize}
	\item We propose to view the MLC task as a sequence generation problem to take the correlations between labels into account.
    \item We propose a sequence generation model with a novel decoder structure, which not only captures the correlations between labels, but also selects the most informative words automatically when predicting different labels.
    \item Extensive experimental results show that our proposed methods outperform the baselines by a large margin. Further analysis demonstrates the effectiveness of the proposed methods on correlation representation.
\end{itemize}

The whole paper is organized as follows. We describe our methods in Section~\ref{methods}. In Section~\ref{experiments}, we present the experiments and make analysis and discussions. Section~\ref{related} introduces the related work. Finally in Section~\ref{conclusion} we conclude this paper and explore the future work.

\section{Proposed Method}\label{methods}
We introduce our proposed methods in detail in this section. 
First, we give an overview of the model in Section~\ref{overview}. Second, we explain the details of the proposed sequence generation model in Section~\ref{seq2seq}. Finally, Section~\ref{global} presents our novel decoder structure.

\subsection{Overview}\label{overview}
First of all, we define some notations and describe the MLC task. Given the label space with $L$ labels $\mathcal{L} = \{l_1, l_2, \cdots, l_L\}$, a text sequence $\bm{x}$ containing $m$ words, the task is to assign a subset $\bm{y}$ containing $n$ labels in the label space $\mathcal{L}$ to $\bm{x}$. Unlike traditional single-label classification where only one label is assigned to each sample, each sample in the MLC task can have multiple labels. From the perspective of sequence generation, the MLC task can be modeled as finding an optimal label sequence $\bm{y^{\ast}}$ that maximizes the conditional probability $p(\bm{y} | \bm{x})$, which is calculated as follows:
\begin{equation}\label{equ1}
p(\bm{y} | \bm{x}) = \prod_{i=1}^{n} p(y_i|y_1, y_2, \cdots, y_{i-1}, \bm{x})
\end{equation}

\begin{figure}[tb]
	\centering
	\includegraphics[width=0.9\linewidth]{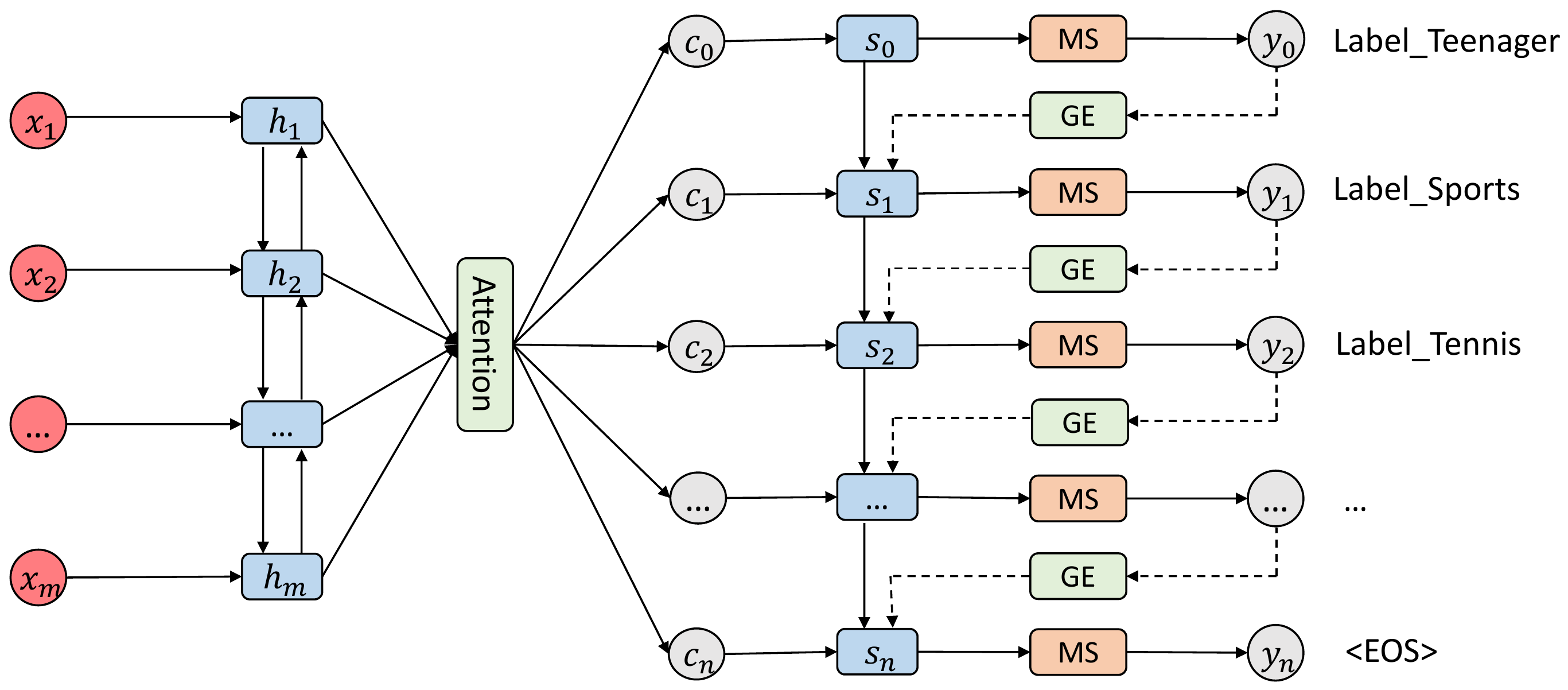}
	\caption{The overview of our proposed model. MS denotes the masked softmax layer. GE denotes the global embedding.}\label{model_fig}
\end{figure}

An overview of our proposed model is shown in Figure \ref{model_fig}. First, we sort the label sequence of each sample according to the frequency of the labels in the training set. High-frequency labels are placed in the front. In addition, the $bos$ and $eos$ symbols are added to the head and tail of the label sequence, respectively. 

The text sequence $\bm{x}$ is encoded to the the hidden states, which are aggregated to a context vector $\bm{c}_t$ by the attention mechanism at time-step $t$. The decoder takes the context vector $\bm{c}_t$, the last hidden state $\bm{s}_{t-1}$ of the decoder and the embedding vector $g(\bm{y}_{t-1})$ as the inputs to produce the hidden state $\bm{s}_t$ at time-step $t$. Here $\bm{y}_{t-1}$ is the predicted probability distribution over the label space $\mathcal{L}$ at time-step $t-1$. The function $g$ takes $\bm{y}_{t-1}$ as input and produces the embedding vector which is then passed to the decoder. Finally, the masked softmax layer is used to output the probability distribution $\bm{y}_t$.

\subsection{Sequence Generation}\label{seq2seq}
In this subsection, we introduce the details of our proposed model. 
The whole sequence generation model consists of an encoder and a decoder with the attention mechanism.

\textbf{Encoder:} Let $(\bm{w}_1, \bm{w}_2, \cdots, \bm{w}_m)$ be a sentence with $m$ words and $\bm{w}_i$ is the one-hot representation of the $i$-th word. We first embed $\bm{w}_i$ to a dense embedding vector $\bm{x}_i$ by an embedding matrix $\bm{E} \in \mathbb{R}^{k \times |\mathcal{V}|}$. Here $|\mathcal{V}|$ is the size of the vocabulary, and $k$ is the dimension of the embedding vector.

We use a bidirectional LSTM \cite{lstm} to read the text sequence $\bm{x}$ from both directions and compute the hidden states for each word,
\begin{align}
\overrightarrow{\bm{h}}_i & = \overrightarrow{\rm LSTM}(\overrightarrow{\bm{h}}_{i-1}, \bm{x}_i) \label{equ2} \\
\overleftarrow{\bm{h}}_i & = \overleftarrow{\rm LSTM}(\overleftarrow{\bm{h}}_{i+1}, \bm{x}_i) \label{equ3}
\end{align}

We obtain the final hidden representation of the $i$-th word by concatenating the hidden states from both directions, $\bm{h}_i = [\overrightarrow{\bm{h}}_i; \overleftarrow{\bm{h}}_i]$, which embodies the information of the sequence centered around the $i$-th word.

\textbf{Attention:} When the model predicts different labels,  not all text words make the same contribution. The attention mechanism produces a context vector by focusing on different portions of the text sequence and aggregating the hidden representations of those informative words. Specially, the attention mechanism assigns the weight $\alpha_{ti}$ to the $i$-th word at time-step $t$ as follows:
\begin{align}
e_{ti} & = \bm{v}_a^T\tanh(\bm{W}_a\bm{s}_t + \bm{U}_a\bm{h}_i) \label{equ3} \\
\alpha_{ti} & = \frac{\exp(e_{ti})}{\sum_{j=1}^{m}\exp(e_{tj})} \label{equ4}
\end{align}
where $\bm{W}_a$, $\bm{U}_a$, $\bm{v}_a$ are weight parameters and $\bm{s}_t$ is the current hidden state of the decoder at time-step $t$. For simplicity, all bias terms are omitted in this paper. The final context vector $\bm{c}_t$ which is passed to the decoder at time-step $t$ is calculated as follows:
\begin{equation}\label{equ5}
\bm{c}_t = \sum_{i=1}^{m}\alpha_{ti}\bm{h}_i
\end{equation}

\textbf{Decoder:} The hidden state $\bm{s}_t$ of the decoder at time-step $t$ is computed as follows:
\begin{equation}\label{d1}
\bm{s}_t = {\rm LSTM}(\bm{s}_{t-1},\left[g(\bm{y}_{t-1}); \bm{c}_{t-1} \right])
\end{equation}
where $[g(\bm{y}_{t-1}); \bm{c}_{t-1}]$ means the concatenation of the vectors $g(\bm{y}_{t-1})$ and $\bm{c}_{t-1}$. $g(\bm{y}_{t-1})$ is the embedding of the label which has the highest probability under the distribution $\bm{y}_{t-1}$.
$\bm{y}_{t-1}$ is the probability distribution over the label space $\mathcal{L}$ at time-step $t-1$ and is computed as follows:
\begin{align}
\bm{o}_t & = \bm{W}_of(\bm{W}_d\bm{s}_t + \bm{V}_d\bm{c}_t) \label{d2} \\
\bm{y}_t & = softmax(\bm{o}_t + \bm{I}_t) \label{d3}
\end{align}
where $\bm{W}_o$, $\bm{W}_d$, and $\bm{V}_d$ are weight parameters, $\bm{I}_t \in \mathbb{R}^L$ is the mask vector that is used to prevent the decoder from predicting repeated labels, and $f$ is a nonlinear activation function.
\begin{equation}\label{d4}
(\bm{I}_t)_i=
\begin{cases}
- \infty & \text{if the label $l_i$ has been predicted at previous $t-1$ time steps.}\\
0 & \text{otherwise.}
\end{cases}
\end{equation}

At the training stage, the loss function is the cross-entropy loss function. We employ the beam search algorithm \cite{beamsearch} to find the top-ranked prediction path at inference time. The prediction paths ending with the $eos$ are added to the candidate path set. 

\subsection{Global Embedding}\label{global}
In the sequence generation model mentioned above, the embedding vector $g(\bm{y}_{t-1})$ in Equation~\eqref{d1} is the embedding of the label that has the highest probability under the distribution $\bm{y}_{t-1}$. However, this calculation only takes advantage of the maximum value of $\bm{y}_{t-1}$ greedily. The proposed sequence generation model generates labels sequentially and predicts the next label conditioned on its previously predicted labels. Therefore, it is likely that we would get a succession of wrong label predictions in the following time steps if the prediction is wrong at time-step $t$, which is also called \emph{exposure bias}. To a certain extent, the beam search algorithm alleviates this problem. However, it can not fundamentally solve the problem because the \emph{exposure bias} phenomenon is likely to occur for all candidate paths. $\bm{y}_{t-1}$ represents the predicted probability distribution at time-step $t-1$, so it is obvious that all information in $\bm{y}_{t-1}$ is helpful when we predict the current label at time-step $t$. The \emph{exposure bias} problem ought to be relieved by considering all informative signals contained in $\bm{y}_{t-1}$.

Based on this motivation, we propose a new decoder structure, where the embedding vector $g(\bm{y}_{t-1})$ at time-step $t$ is capable of representing the overall information at $(t-1)$-th time step. Inspired by the idea of the adaptive gate in highway network \cite{highway}, here we introduce our global embedding. Let $\bm{e}$ denotes the embedding of the label which has the highest probability under the distribution $\bm{y}_{t-1}$. $\bm{\bar{e}}$ is the weighted average embedding at time $t$, which is calculated as follows:
\begin{equation}\label{g1}
\bm{\bar{e}} = \sum_{i=1}^{L}y_{t-1}^{(i)}\bm{e}_i
\end{equation}
where $y_{t-1}^{(i)}$ is the $i$-th element of $\bm{y}_{t-1}$ and $\bm{e}_i$ is the embedding vector of the $i$-th label. Then the proposed global embedding $g(\bm{y}_{t-1})$ passed to the decoder at time-step $t$ is as follows:
\begin{equation}\label{g2}
g(\bm{y}_{t-1}) = (\bm{1}-\bm{H}) \odot \bm{e} + \bm{H} \odot \bm{\bar{e}}
\end{equation}
where $\bm{H}$ is the transform gate controlling the proportion of the weighted average embedding:
\begin{equation}\label{g3}
\bm{H} = \bm{W}_1\bm{e} + \bm{W}_2\bm{\bar{e}}
\end{equation}
where $\bm{W}_1, \bm{W}_2 \in \mathbb{R}^{L \times L}$ are weight matrices. The global embedding $g(\bm{y}_{t-1})$ is the optimized combination of the original embedding and the weighted average embedding by using transform gate $\bm{H}$, which can automatically determine the combination factor in each dimension. $\bm{y}_{t-1}$ contains the information of all possible labels. By considering the probability of every label, the model is capable of reducing damage caused by mispredictions made in the previous time steps. This enables the model to predict label sequences more accurately.

\section{Experiments}\label{experiments}
In this section, we evaluate our proposed methods on two datasets. We first introduce the datasets, evaluation metrics, experimental details, and all baselines. Then, we compare our methods with the baselines. Finally, we provide the analysis and discussions of experimental results.

\subsection{Datasets}

\noindent\textbf{Reuters Corpus Volume I (RCV1-V2)\footnote{\url{http://www.ai.mit.edu/projects/jmlr/papers/volume5/lewis04a/lyrl2004_rcv1v2_README.htm}}:}  
This dataset is provided by \newcite{rcv1}. It consists of over \num{800000} manually categorized newswire stories made available by Reuters Ltd for research purposes. Multiple topics can be assigned to each newswire story and there are 103 topics in total. 

\noindent\textbf{Arxiv Academic Paper Dataset (AAPD)\footnote{\url{https://github.com/lancopku/SGM}}:} We build a new large dataset for the multi-label text classification. We collect the abstract and the corresponding subjects of \num{55840} papers in the computer science field from the website\footnote{\url{https://arxiv.org/}}. An academic paper may have multiple subjects and there are 54 subjects in total. The target is to predict corresponding subjects of an academic paper according to the content of the abstract. 

We divide each dataset into training, validation and test sets. The statistics of the two datasets are shown in Table~\ref{tab_datasets}.

\begin{table}[t]
	\centering
    \footnotesize
    \setlength{\tabcolsep}{12pt}
	\begin{tabular}{|l|c|c|c|c|}
		\hline
        \multicolumn{1}{|l|}{\textbf{Dataset}} &
		\multicolumn{1}{|c}{\textbf{Total Samples}} &
        \multicolumn{1}{|c}{\textbf{Label Sets}} &
        \multicolumn{1}{|c}{\textbf{Words/Sample}} &
		\multicolumn{1}{|c|}{\textbf{Labels/Sample}}   \\ \hline
        RCV1-V2 & \num{804414}  & 103 & 123.94 & 3.24 \\ \hline
  		AAPD & \num{55840} & \num{54} & \num{163.42} & \num{2.41} \\ \hline
	\end{tabular}
  	\caption{Summary of datasets. \textbf{Total Samples}, \textbf{Label Sets} denote the total number of samples and labels, respectively. \textbf{Words/Sample} is the average number of words per sample and \textbf{Labels/Sample} is the average number of labels per sample.}\label{tab_datasets}
    \vspace{-0.1in}
\end{table}

\subsection{Evaluation Metrics}
Following the previous work~\cite{ml_6,r6}, we adopt hamming loss and micro-$F_1$ score as our main evaluation metrics. Micro-precision and micro-recall are also reported to assist the analysis.

\begin{itemize}

	\item \noindent\textbf{Hamming-loss} \cite{hamming_loss} evaluates the fraction of misclassified instance-label pairs, where a relevant label is missed or an irrelevant is predicted. 

	\item \noindent\textbf{Micro-$F_1$} \cite{f1} can be interpreted as a weighted average of the precision and recall. It is calculated globally by counting the total true positives, false negatives, and false positives. 
    
\end{itemize}

\subsection{Details}



We extract the vocabularies from the training sets. For the RCV1-V2 dataset, the size of the vocabulary is \num{50000} and out-of-vocabulary (OOV) words are replaced with $unk$. Each document is truncated at the length of 500 and the beam size is 5 at the inference stage. Besides, we set the word embedding size to 512. The hidden sizes of the encoder and the decoder are 256 and 512, respectively. The number of LSTM layers of encoder and decoder is 2.

For the AAPD dataset, the size of word embedding is 256. There are two LSTM layers in the encoder and its size is 256. For the decoder, there is one LSTM layer of size 512. The size of the vocabulary is \num{30000} and OOV words are also replaced with $unk$. Each document is truncated at the length of 500. The beam size is 9 at the inference stage.

We use the Adam~\cite{KingmaBa2014} optimization method to minimize the cross-entropy loss over the training data. For the hyper-parameters of the Adam optimizer, we set the learning rate $\alpha = 0.001$, two momentum parameters $\beta_{1} = 0.9$ and $\beta_{2} = 0.999$ respectively, and $\epsilon = 1 \times 10^{-8}$. Additionally, we make use of the dropout regularization \cite{dropout} to avoid overfitting and clip the gradients~\cite{gradientclip} to the maximum norm of 10.0. During training, we train the model for a fixed number of epochs and monitor its performance on the validation set. Once the training is finished, we select the model with the best micro-$ F_1$ score on the validation set as our final model and evaluate its performance on the test set.

\subsection{Baselines}
We compare our proposed methods with the following baselines:

\begin{itemize}
	
	\item \noindent\textbf{Binary Relevance (BR)}~\cite{r1} transforms the MLC task into multiple single-label classification problems by ignoring the correlations between labels.
	
	\item \noindent\textbf{Classifier Chains (CC)}~\cite{ml_3} transforms the MLC task into a chain of binary classification problems and takes high-order label correlations into consideration.
    
    \item \noindent\textbf{Label Powerset (LP)}~\cite{lp} transforms a multi-label problem to a multi-class problem with one multi-class classifier trained on all unique label combinations.
	
	
    
    \item \noindent\textbf{CNN}~\cite{cnn} uses multiple convolution kernels to extract text features, which are then inputted to the linear transformation layer followed by a sigmoid function to output the probability distribution over the label space. The multi-label soft margin loss is optimized.
	
	\item \noindent\textbf{CNN-RNN}~\cite{r6} utilizes CNN and RNN to capture both the global and local textual semantics and model the label correlations.

\end{itemize}

Following the previous work~\cite{r6}, we adopt the linear SVM as the base classifier in BR, CC and LP. We implement BR, CC and LP by means of Scikit-Multilearn \cite{sklearn}, an open-source library for the MLC task.
We tune hyper-parameters of all baseline algorithms on the validation set based on the micro-$F_1$ score. In addition, training strategies mentioned in \newcite{cnn_trick} are used to tune hyper-parameters for the baselines CNN and CNN-RNN.

\subsection{Results}
For the purpose of simplicity, we denote the proposed sequence generation model as \textbf{SGM}. We report the evaluation results of our methods and all baselines on the test sets. 
\begin{table}[tb]
	\begin{subtable}[h]{0.5\textwidth}
		\centering
        \footnotesize
    	\setlength{\tabcolsep}{9.5pt}
		\begin{tabular}{| l | c c c c |}
		\hline
		\multicolumn{1}{| l |}{\textbf{Models}} &
		\multicolumn{1}{c}{\textbf{HL(-)}} & 
		\multicolumn{1}{c}{\textbf{P(+)}} &  
        \multicolumn{1}{c}{\textbf{R(+)}} &
		\multicolumn{1}{c |}{\textbf{F1(+)}}   \\  \hline
        BR & 0.0086 & 0.904 & 0.816 & 0.858\\
		CC & 0.0087  & 0.887 & 0.828 & 0.857\\
        LP & 0.0087  & 0.896 & 0.824 & 0.858 \\
        CNN & 0.0089  & \textbf{0.922} & 0.798 & 0.855 \\
		CNN-RNN & 0.0085 & 0.889 & 0.825 & 0.856 \\ \hline
        SGM & 0.0081  & 0.887 & 0.850 & 0.869 \\ 
        + GE & \textbf{0.0075} & 0.897 & \textbf{0.860} & \textbf{0.878}  \\ \hline
		\end{tabular}
		\caption{Performance on the RCV1-V2 test set.}
		\label{tab_rcv1}
	\end{subtable}
	\hfill
	\begin{subtable}[h]{0.5\textwidth}
		\centering
        \footnotesize
    	\setlength{\tabcolsep}{9.5pt}
		\begin{tabular}{| l | c c c c |}
		\hline
		\multicolumn{1}{| l |}{\textbf{Models}} &
		\multicolumn{1}{c}{\textbf{HL(-)}} & 
		\multicolumn{1}{c}{\textbf{P(+)}} &  
        \multicolumn{1}{c}{\textbf{R(+)}} &
		\multicolumn{1}{c |}{\textbf{F1(+)}}   \\ \hline
        BR & 0.0316 & 0.644 & 0.648 & 0.646 \\
		CC & 0.0306 & 0.657 & 0.651 & 0.654 \\
        LP & 0.0312  & 0.662 & 0.608 & 0.634 \\
        CNN & 0.0256 & \textbf{0.849} & 0.545 & 0.664 \\
		CNN-RNN & 0.0278 & 0.718 & 0.618 & 0.664 \\ \hline
        SGM & 0.0251 & 0.746 & 0.659 & 0.699 \\ 
        + GE & \textbf{0.0245} & 0.748 & \textbf{0.675} & \textbf{0.710} \\ \hline
		\end{tabular}
		\caption{Performance on the AAPD test set.}
		\label{tab_arxiv_result}
	\end{subtable}
	\caption{Comparison between our methods and all baselines on two datasets. GE denotes the global embedding. HL, P, R, and F1 denote hamming loss, micro-precision, micro-recall, and micro-$F_1$, respectively. The symbol “+” indicates that the higher the value is, the better the model performs. The symbol “-” is the opposite.}
	\label{tab_result}
    \vspace{-0.1in}
\end{table}

The experimental results of our methods and the baselines on dataset RCV1-V2 are shown in Table~\ref{tab_rcv1}. Results show that our proposed methods give the best performance in the main evaluation metrics. Our proposed SGM model using global embedding achieves a reduction of 12.79\% hamming-loss and an improvement of 2.33\% micro-$F_1$ score over the most commonly used baseline BR. Besides, our methods outperform other traditional deep-learning models by a large margin. For instance, the proposed SGM model with global embedding achieves a reduction of 15.73\% hamming-loss and an improvement of 2.69\% micro-$F_1$ score over the traditional CNN model. Even without the global embedding, our proposed SGM model is still able to outperform all baselines. 

In addition, the SGM model is significantly improved by using global embedding. The SGM model with global embedding achieves a reduction of 7.41\% hamming loss and an improvement of 1.04\% micro-$F_1$ score on the test set compared with the model without global embedding. 

Table~\ref{tab_arxiv_result} presents the results of the proposed methods and the baselines on the AAPD test set. Similar to the experimental results on the RCV1-V2 test set, our proposed methods still outperform all baselines by a large margin in main evaluation metrics. This further confirms that our methods have significant advantages over previous work on large datasets. Besides, the proposed SGM achieves a reduction of 2.39\% hamming loss and an improvement of 1.57\% micro-$F_1$ score on the test set by using global embedding. This further testifies that the global embedding is capable of helping the model to predict label sequences more accurately. 


\subsection{Analysis and Discussion}
Here we perform further analysis on the model and experimental results. We report the evaluation results in terms of hamming loss and micro-$F_1$ score.

\subsubsection{Exploration of Global Embedding}

\begin{figure}
\begin{minipage}[t]{0.495\textwidth}
\centering
\includegraphics[width=1.0\linewidth]{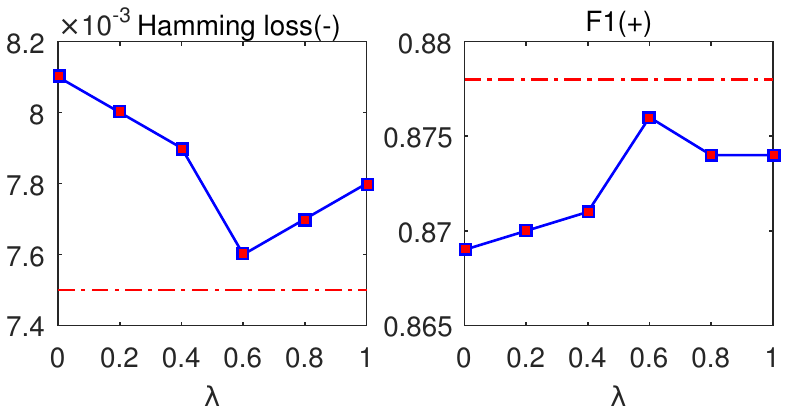}
\caption{The performance of the SGM model when using different $\lambda$. The red dotted line represents the results of using the adaptive gate. The symbol “+” indicates that the higher the value is, the better the model performs. The symbol “-” is the opposite.}
\label{figure_lambda}
\vspace{-0.2in}
\end{minipage}%
\hspace{0.15in}
\begin{minipage}[t]{0.495\textwidth}
\centering
\includegraphics[width=1.0\linewidth]{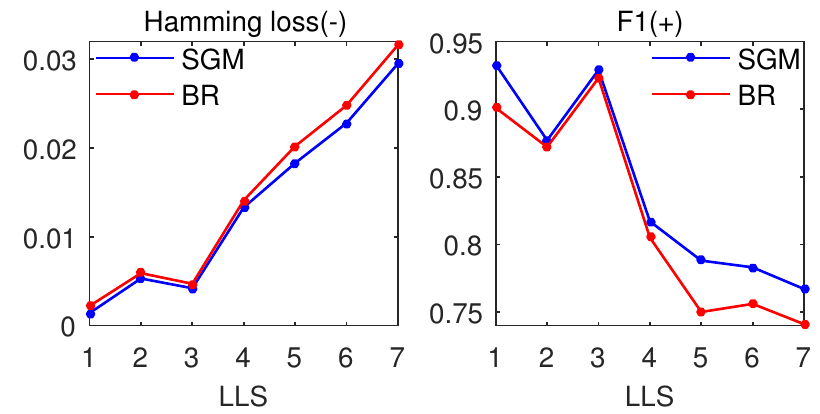}
\caption{The performance of the SGM model on different subsets of the RCV1-V2 test set. LLS represents the length of label sequence of each sample in the subset. The explanations of symbol “+” and “-” can be found in Figure~\ref{figure_lambda}.}
\label{figure_N}
\vspace{-0.1in}
\end{minipage}
\end{figure}

As is shown in Table~\ref{tab_result}, global embedding can significantly improve the performance of the model. The global embedding $g(\bm{y}_{t-1})$ at time-step $t$ takes advantage of all information of possible labels contained in $\bm{y}_{t-1}$, so it is able to enrich the source information when the model predicts the current label, which leads to the performance of the model significantly improved. The global embedding is the combination of original embedding $\bm{e}$ and the weighted average embedding $\bm{\bar{e}}$ by using the transform gate $\bm{H}$. Here we conduct experiments on the RCV1-V2 dataset to explore how the performance of our model is affected by the proportion between two kinds of embeddings. In the exploratory experiment, the final embedding vector at time-step $t$ is calculated as follows:
\begin{equation}\label{lambda}
g(\bm{y}_{t-1}) = (1-\lambda) * \bm{e} + \lambda * \bm{\bar{e}}
\end{equation}
The proportion between two kinds of embeddings is controlled by coefficient $\lambda$. $\lambda = 0$ denotes the proposed SGM model without global embedding. The proportion of weighted average embedding increases when we increase $\lambda$. The experimental results using different $\lambda$ values in the decoder are shown in Figure~\ref{figure_lambda}.

As is shown in Figure~\ref{figure_lambda}, the performance of the model varies when different $\lambda$ is used. Overall, the model using the adaptive gate performs the best, which achieves the best results in both hamming loss and micro-$F_1$.
The models with $\lambda \neq 0$ outperform the model with $\lambda=0$, which shows that the weighted average embedding contains richer information, leading to the improvement in the performance of the model. Without using the adaptive gate, the performance of the model improves at first and then deteriorates as $\lambda$ increases.
It reveals the reason why the model with the adaptive gate performs the best: the adaptive gate can automatically determine the most appropriate $\lambda$ value according to the actual condition. 

\subsubsection{The Impact of Mask and Sorting}
Our proposed methods are developed based on traditional Seq2Seq models. However, the mask module is added to the proposed methods, which is used to prevent the models from predicting repeated labels. In addition, we sort the label sequence of each sample according to the frequency of appearance of labels in the training set. In order to explore the impact of the mask module and sorting,  we conduct ablation experiments on the RCV1-V2 dataset. The experimental results are shown in Table~\ref{tab_ablation}. “\emph{w/o mask}” means that we do not perform mask operation and “\emph{w/o sorting}” means that we randomly shuffle the label sequence in order to perturb its original order.

As is shown in Table~\ref{tab_ablation}, the performance decline of the SGM model with global embedding is more significant compared with that of the SGM model without global embedding. In addition, the decline in the performance of the two models is more significant when we randomly shuffle the label sequence of the sample compared with removing mask module. The label cardinality of the RCV1-V2 dataset is small, so our proposed methods are less prone to predicting repeated labels. This explains the reason why experimental results indicate that the mask module has little impact on the models' performance. In addition, the proposed models are trained using the maximum likelihood estimation method and the cross-entropy loss function, which requires humans to predefine the order of the output labels. Therefore, the sorting of labels is very important for the models' performance. 
Besides, the performance of both models declines when we do not use the mask module. This shows that the performance of the model can be improved by using the mask operation.

\subsubsection{Error Analysis} \label{error analysis}

\begin{table}[tb]
	\begin{subtable}[h]{0.5\textwidth}
		\centering
        \footnotesize
    	\setlength{\tabcolsep}{9.5pt}
		\begin{tabular}{| l | c c |}
		\hline
		\multicolumn{1}{| l |}{\textbf{Models}} &
		\multicolumn{1}{c}{\textbf{HL(-)}} & 
		\multicolumn{1}{c |}{\textbf{F1(+)}}   \\  \hline
         SGM & 0.0081 & 0.869 \\ \hline
  \emph{w/o mask} & 0.0083($\downarrow 2.47\%$) & 0.866($\downarrow 0.35\%$) \\ 
  \emph{w/o sorting} & 0.0084($\downarrow 3.70\%$) & 0.858($\downarrow 1.27\%$) \\ \hline
		\end{tabular}
		\caption{Ablation study for the SGM model.}
		\label{tab_ablation1}
	\end{subtable}
	\hfill
	\begin{subtable}[h]{0.5\textwidth}
		\centering
        \footnotesize
    	\setlength{\tabcolsep}{9.5pt}
		\begin{tabular}{| l | c c |}
		\hline
		\multicolumn{1}{| l |}{\textbf{Models}} &
		\multicolumn{1}{c}{\textbf{HL(-)}} & 
		\multicolumn{1}{c |}{\textbf{F1(+)}}   \\ \hline
         SGM + GE & 0.0075 & 0.878 \\ \hline
  \emph{w/o mask} & 0.0078($\downarrow 4.00\%$) & 0.873($\downarrow 0.57\%$) \\
  \emph{w/o sorting} & 0.0083(\textbf{$\downarrow10.67\%$}) & 0.859($\downarrow 2.16\%$)\\ \hline
		\end{tabular}
		\caption{Ablation study for SGM model with global embedding.}
		\label{tab_ablation2}
	\end{subtable}
	\caption{Ablation study on the RCV1-V2 test set. GE denotes the global embedding. HL and F1 denote hamming loss and micro-$F_1$, respectively. The symbol “+” indicates that the higher the value is, the better the model performs. The symbol “-” is the opposite. $\uparrow$ means that the performance of the model improves and $\downarrow$ is the opposite. }
	\label{tab_ablation}
\end{table}

In the experiment, we find that the performance of all methods deteriorates when the length of the label sequence increases (for simplicity, we denote the length of the label sequence as \emph{LLS}). In order to explore the influence of the value of the \emph{LLS}, we divide the test set into different subsets based on  different \emph{LLS}. Figure~\ref{figure_N} shows the performance of the SGM model and the most commonly used baseline BR on different subsets of the RCV1-V2 test set. As is shown in Figure~\ref{figure_N}, generally, the performance of both models deteriorates as the \emph{LLS} increases. This shows that when the label sequence of the sample is particularly long, it is difficult to accurately predict all labels. Because more information is needed when the model predicts more labels. It is easy to ignore some true labels whose feature information is insufficient.

However, as is shown in Figure~\ref{figure_N}, the proposed SGM model outperforms BR with any value of \emph{LLS}, and the advantages of our model are more significant when \emph{LLS} is large. The traditional BR method predicts all labels at once only based on the sample input. Therefore, it tends to ignore some true labels whose feature information contained in the sample is insufficient. The SGM model generates labels sequentially, and predicts the next label based on its previously predicted labels. Therefore, even if the sample contains less information of some true labels, the SGM model is capable of generating these true labels by considering relevant labels that have been predicted.

\subsubsection{Visualization of Attention}

\begin{table}[tb]
	\begin{subtable}[h]{0.5\textwidth}
		\centering
        \footnotesize
    	\setlength{\tabcolsep}{19pt}
		\begin{tabular}{|p{0.8\textwidth}|}
		\hline
        $\bullet$ \fancyhl{2}{Generating} descriptions for \fancyhl{42}{videos} has many 
        applications including human \fancyhl{37}{robot} interaction. \\ \hline
        $\bullet$ Many methods for \fancyhl{100}{image }\fancyhl{61}{captioning} rely on 
        pre-trained \fancyhl{45}{object }\fancyhl{12}{classifier }\fancyhl{51}{CNN} and Long 
        Short Term \fancyhl{6}{Memory} recurrent networks. \\ \hline
        $\bullet$ How to learn \fancyhl{21}{robust }\fancyhl{61}{visual }\fancyhl{33}{classifiers} from 
 		the weak annotations of the sentence descriptions.\\ \hline
		\end{tabular}
		\caption{Visual analysis when the SGM model predicts “\emph{CV}”.}
		\label{tab_av1}
	\end{subtable}
	\hfill
	\begin{subtable}[h]{0.5\textwidth}
		\centering
        \footnotesize
    	\setlength{\tabcolsep}{19pt}
		\begin{tabular}{|p{0.8\textwidth}|}
		\hline
        $\bullet$ \fancyhl{10}{Generating} descriptions for videos has many 
        applications including human robot interaction. \\ \hline
        $\bullet$ Many methods for \fancyhl{6}{image }\fancyhl{8}{captioning} rely on 
        pre-trained object \fancyhl{10}{classifier} CNN and \fancyhl{37}{Long}  
        Short Term \fancyhl{57}{Memory }\fancyhl{65}{recurrent} networks. \\ \hline
        $\bullet$ How to learn \fancyhl{29}{robust} visual \fancyhl{33}{classifiers} from the 
 		weak \fancyhl{37}{annotations} of the \fancyhl{100}{sentence} descriptions.\\ \hline
		\end{tabular}
		\caption{Visual analysis when the SGM model predicts “\emph{CL}”.}
		\label{tab_visual1}
	\end{subtable}
	\caption{An example abstract in the AAPD dataset, from which we extract three informative sentences. This abstract is assigned two labels: “\emph{CV}” and “\emph{CL}”. They denote computer vision and computational language, respectively.}
	\label{tab_visual}
\end{table}

When the model predicts different labels, there exist differences in the contributions of different words. The SGM model is able to select the most informative words by utilizing the attention mechanism. The visualization of the attention layer is shown in Table~\ref{tab_visual}. According to Table~\ref{tab_visual}, when the SGM model predicts the label “\emph{CV}”, it can automatically assign larger weights to more informative words, like \texttt{image}, \texttt{visual}, \texttt{captioning}, and so on. For the label “\emph{CL}”, the selected informative words are \texttt{sentence}, \texttt{memory}, \texttt{recurrent}, etc. This shows that our proposed models are able to consider the differences in the contributions of textual content when predicting different labels and select the most informative words automatically.

\subsubsection{Case Study}
We give several examples of the generated label sequences on the RCV1-V2 dataset in Table~\ref{case}, where we compare the proposed methods with the most commonly used baseline BR. The red bold labels in each example indicate that they are highly correlated. For instance, the correlation coefficient between E51 and E512 is 0.7664. Therefore, these highly correlated labels are likely to appear together in the predicted label sequence. The BR algorithm fails to capture this label correlation, leaving many true labels unpredicted. However, our proposed methods accurately predict almost all highly correlated true labels. The proposed SGM captures the correlations between labels by utilizing LSTM to generate labels sequentially. Therefore, for some true labels whose feature information is insufficient, the proposed SGM is still able to generate them by considering relevant labels that have been predicted.
In addition, label sequences that are more accurate are predicted by using global embedding. The SGM model with global embedding predicts more true labels compared with the SGM model without global embedding. The reason is that the source information is further enriched by incorporating overall informative signals in the probability distribution $\bm{y}_{t-1}$ when the model predicts the label at time-step $t$. Enriched information makes global embedding more smooth, which enables the model to reduce damage caused by mispredictions made in the previous time steps.

\section{Related Work}\label{related}

\begin{table}[t]
	\centering
    \footnotesize
    \setlength{\tabcolsep}{1.5pt}
	\begin{tabular}{|p{0.24\linewidth}|p{0.24\linewidth}|p{0.24\linewidth}|p{0.24\linewidth}|}
		\hline
		\textbf{Reference} &
		\textbf{BR} & 
        \textbf{SGM} &
		\textbf{SGM + GE} \\ \hline
        CCAT, {\color{red}{\textbf{C15, C152}}}, C41, C411 & CCAT, C15, C13 & CCAT, {\color{red}{\textbf{C15, C152}}} & CCAT, {\color{red}{\textbf{C15, C152}}}, C41, C411 \\ \hline 
        CCAT, GCAT, ECAT, C31, GDIP, C13, C21, {\color{red}{\textbf{E51, E512}}} & CCAT, GCAT, GDIP, E51 & CCAT, ECAT, GDIP, {\color{red}{\textbf{E51, E512}}} & CCAT, GCAT, ECAT, C31, GDIP, {\color{red}{\textbf{E51, E512}}}, C312 \\ \hline 
        GCAT, ECAT, {\color{red}{\textbf{G15, G154, G151, G155}}} & GCAT, ECAT, GENV, G15 & GCAT, ECAT, E21, {\color{red}{\textbf{G15, G154, G156}}} & GCAT, ECAT, E21, {\color{red}{\textbf{G15, G154, G155}}} \\ \hline 

	\end{tabular}
	\caption{Several examples of the generated label sequences on the RCV1-V2 dataset. The red bold labels in each example indicate that they are highly correlated.}
    \label{case}
\end{table}

The MLC task studies the problem where multiple labels are assigned to each sample. There are four main types of methods for the MLC task: problem transformation methods, algorithm adaptation methods, ensemble methods, and neural network models.

Problem transformation methods map the MLC task into multiple single-label learning tasks. Binary relevance (BR) \cite{r1} decomposes the MLC task into independent binary classification problems by ignoring the correlations between labels. In order to model label correlations, label powerset (LP)~\cite{lp} transforms a multi-label problem to a multi-class problem with a classifier trained on all unique label combinations. Classifier chains (CC) \cite{ml_3} transforms the MLC task into a chain of binary classification problems, where subsequent binary classifiers in the chain are built upon the predictions of preceding ones. However, the computational efficiency and performance of these methods are challenged by applications with a large number of labels and samples.

Algorithm adaptation methods extend specific learning algorithms to handle multi-label data directly. \newcite{ml_7} construct decision tree based on multi-label entropy to perform classification. \newcite{ml_8} optimize the empirical ranking loss by using maximum margin strategy and kernel tricks. Collective multi-label classifier (CML) \cite{cml} adopts maximum entropy principle to deal with multi-label data by encoding label correlations as constraint conditions. \newcite{ml_6} adopt $k$-nearest neighbor techniques to deal with multi-label data. \newcite{ml_5} make ranking among labels by utilizing pairwise comparison. \newcite{li2015multi} propose a novel joint learning algorithm that allows the feedbacks to be propagated from the classifiers for latter labels to the classifier for the current label. Most methods, however, can only be used to capture the first or second order label correlations or are computationally intractable in considering high-order label correlations. 

Among ensemble methods, \newcite{rakel} break the initial set of labels into a number of small random subsets and employ the LP algorithm to train a corresponding classifier. \newcite{lsps} propose to construct a label co-occurrence graph and perform community detection to partition the label set. 

In recent years, some neural network models have also been used for the MLC task. \newcite{r5} propose the BP-MLL that utilizes a fully-connected neural network and a pairwise ranking loss function. \newcite{r3} propose a neural network using cross-entropy loss instead of ranking loss. \newcite{r8} increase classification speed by adding an extra ART layer for clustering. \newcite{r4} utilize word embeddings based on CNN to capture label correlations. \newcite{r6} propose to represent semantic information of text and model high-order label correlations by combining CNN with RNN. \newcite{r7} initialize the final hidden layer with rows that map to co-occurrence of labels based on the CNN architecture to improve the performance of the model. \newcite{ma2018bag} propose to use the multi-label classification algorithm for machine translation to handle the situation where a sentence can be translated into more than one correct sentences.

\section{Conclusions and Future Work}\label{conclusion}
In this paper, we propose to view the multi-label classification task as a sequence generation problem to model the correlations between labels. A sequence generation model with a novel decoder structure is proposed to improve the performance of classification. Extensive experimental results show that the proposed methods outperform the baselines by a substantial margin. Further analysis of experimental results demonstrates that our proposed methods not only capture the correlations between labels, but also select the most informative words automatically when predicting different labels.

As analyzed in Section~\ref{error analysis}, when a large number of labels are assigned to a sample, how to predict all these true labels accurately is an intractable problem. Our proposed methods alleviate this problem to some extent, but more effective solutions need to be further explored in the future.

\section{Acknowledgements}
This work is supported in part by National Natural Science Foundation of China (No. 61673028, No. 61333018) and the National Thousand Young Talents Program. Xu Sun is the corresponding author of this paper.

\bibliographystyle{acl}
\bibliography{acl2018}

\end{document}